\documentclass[10pt, a4paper]{article}

\usepackage{lrec-coling2024} 

\usepackage{booktabs}

\usepackage{amsmath}
\DeclareMathOperator*{\argmax}{arg\,max}

\title{\includegraphics[scale=0.04]{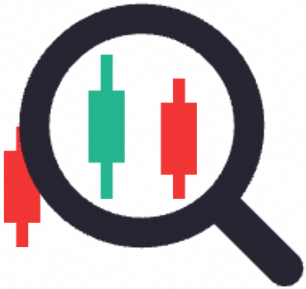} \textbf{\emph{AlphaFin}: Benchmarking Financial Analysis with Retrieval- Augmented Stock-Chain Framework}}

\name{Xiang Li\textsuperscript{1}*, Zhenyu Li\textsuperscript{1}*, Chen Shi\textsuperscript{2}*, \\ \textbf{\large Yong Xu\textsuperscript{1}, Qing Du\textsuperscript{1}$^\dag$, Mingkui Tan\textsuperscript{1}, Jun Huang\textsuperscript{2}, Wei Lin\textsuperscript{2}} \thanks{* Equal contribution. This work was conducted when Xiang Li and Zhenyu Li were interning at Alibaba.}\thanks{$^\dag$ Corresponding author.}} 

\address{\textsuperscript{1}South China University of Technology, China \\ \textsuperscript{2}Alibaba Group, China \\
         \textrm{\{lixiangjacky, zhenyuli2148\}@gmail.com,  deling.sc@alibaba-inc.com}\\
         }

\abstract{
The task of financial analysis primarily encompasses two key areas: stock trend prediction and the corresponding financial question answering.
Currently, machine learning and deep learning algorithms (ML\&DL) have been widely applied for stock trend predictions, leading to significant progress. 
However, these methods fail to provide reasons for predictions, lacking interpretability and reasoning processes. Also, they can not integrate textual information such as financial news or reports.
Meanwhile, large language models (LLMs) have remarkable textual understanding and generation ability. But due to the scarcity of financial training datasets and limited integration with real-time knowledge, LLMs still suffer from hallucinations and are unable to keep up with the latest information. 
To tackle these challenges, 
we first release AlphaFin datasets, combining traditional research datasets, real-time financial data, and handwritten chain-of-thought (CoT) data. It has a positive impact on training LLMs for completing financial analysis.
We then use AlphaFin datasets to benchmark a state-of-the-art method, called Stock-Chain, for effectively tackling the financial analysis task, which integrates retrieval-augmented generation (RAG) techniques.
Extensive experiments are conducted to demonstrate the effectiveness of our framework on financial analysis.
\\ \newline  \\}

\begin{document}

\maketitleabstract

\section{Introduction}

With the advancement of the financial industry, the importance of financial analysis has become increasingly prominent. The ability of financial analysis is primarily manifested in the areas of stock trend prediction and the corresponding financial Q\&A. 
The advent of LLMs has attracted attention from the financial industry, as they possess exceptional generation capabilities~\cite{dredze2016twitter,araci2019finbert,bao2021plato}. Thus, there is a strong desire to leverage these LLMs to enhance the accuracy of financial analysis.\footnote{Resources are publicly available at: \url{https://github.com/AlphaFin-proj/AlphaFin}}
\begin{figure}[t]
\centering
\includegraphics[width=0.5\textwidth, keepaspectratio]{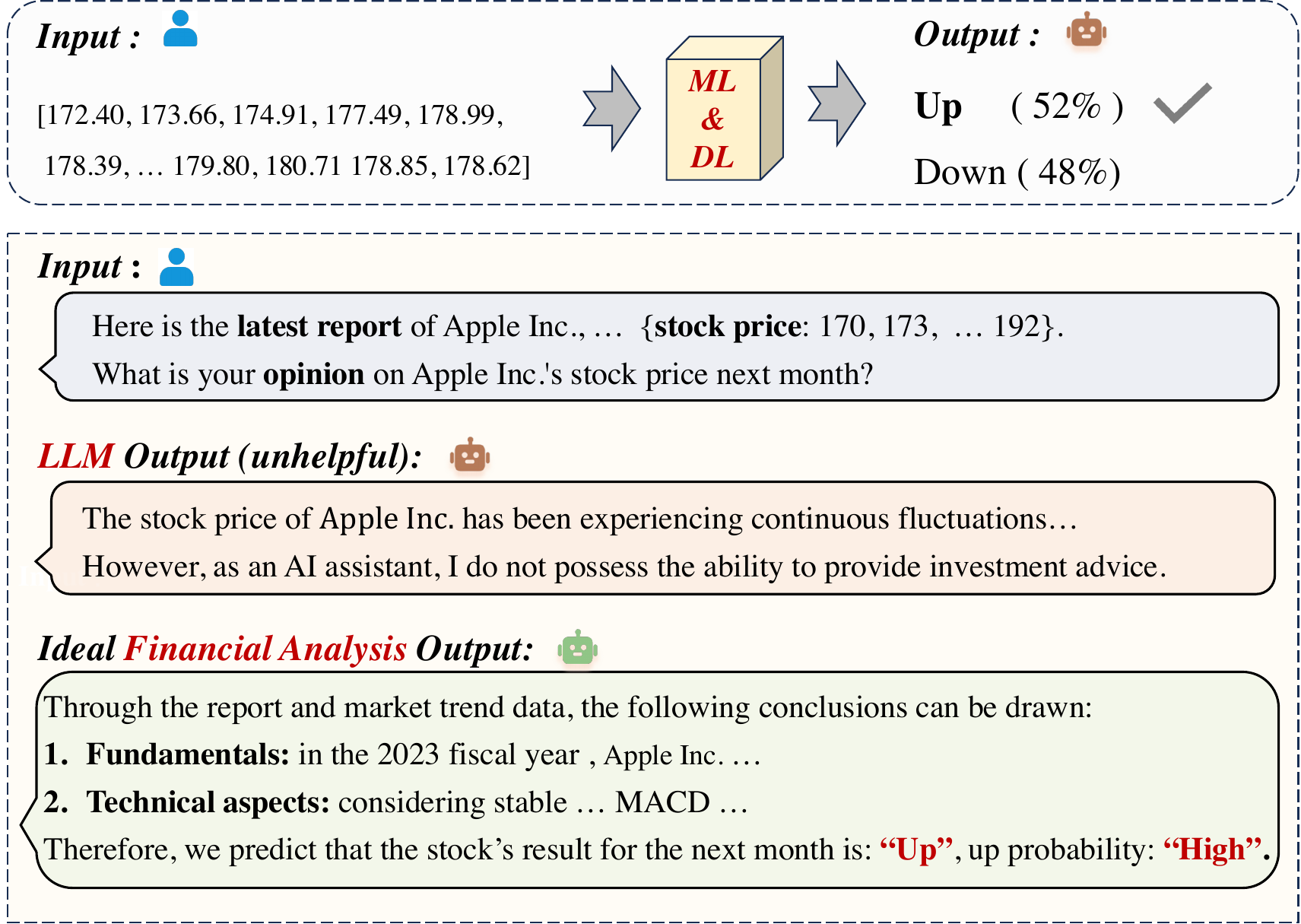}
\caption{An example of the financial analysis task, including stock trend prediction and financial Q\&A. Traditional ML\&DL methods merely provide uncertain forecasts (Up/Down) without any justification, while original LLMs could offer analysis of the prediction but unhelpful.}
\vspace{-1em}
\label{fig1}
\end{figure}

Numerous recent studies have attempted to create efficient algorithms using ML\&DL for stock trend prediction~\cite{saad1998comparative,shah2022comprehensive}.
At present, ML\&DL has been widely used for stock trend prediction based on time series data, generating a positive impact on the industry.
However, ML\&DL algorithms have limited performance, only able to provide uncertain results and incapable of handling complex textual data. Meanwhile, they fail to provide investors with effective justifications and analyze the underlying causes, potentially disrupting their investment confidence. 
As shown in Figure~\ref{fig1}, for Apple Inc., ML\&DL (such as LSTM) can predict an uncertain stock trend (``up'') for the next month based on the former stock price data.
Nevertheless, it could not provide a reliable result and analysis of the prediction. 
If they could offer effective analysis, it would greatly enhance investor's confidence in decision-making. 

Fortunately, LLMs possess excellent capabilities in text processing and generation.
To utilize the capacity of LLMs, FinGPT~\cite{yang2023fingpt} and BloombergGPT~\cite{wu2023bloomberggpt} have been specifically designed as FinLLMs. They can be applied to various financial tasks, catering to the needs of the industry. As shown in Figure~\ref{fig1}, they have the potential to handle diverse text data, including news and reports~\cite{zhang2023instructfingpt}. This advancement empowers investors to make precise investment and trading decisions. 

Nevertheless, building a FinLLMs is not a straightforward task. 
LLMs often exhibit phenomena such as hallucination and meaningless outputs~\cite{mundler2023self}, even for the advanced ChatGPT~\cite{ouyang2022training}.
As shown in Figure 1, the content generated by general LLMs lacks helpfulness and fails to meet real-time requirements. This can be attributed to two reasons. 
Firstly, the quality of generated content relies on data availability. The absence of high-quality financial training datasets~\cite{radford2019language} impacts the quality of generation.
Secondly, stock trend prediction relies on precise and real-time information, the absence of these information leads to LLMs hallucinations. 
Despite the widespread application of RAG~\cite{lewis2021retrievalaugmented} in other fields, their adaptation in the financial domain remains insufficient.




To tackle these challenges, in this paper, we formalize the task of financial analysis and release AlphaFin for fine-tuning FinLLMs, which contain traditional research datasets, real-time financial data, and handwritten CoT data.  Moreover, we propose a Stock-Chain framework integrated with RAG. Stock-Chain not only provides investors with stock trend prediction but also integrates real-time market data and macroeconomic news through RAG, enabling accurate stock analysis during interactions with investors.

Experimental results demonstrate that Stock-Chain is able to achieve the task of stock trend prediction with state-of-the-art accuracy and over 30\% annualized rate of return (ARR). 
Meanwhile, Stock-Chain can provide comprehensive analysis in the financial Q\&A, enhancing investors' confidence in decision-making and providing a solid foundation for their investment choices. 
We conduct extensive supplementary experiments like ablation study, GPT4\&human preference evaluation, and case study.

In summary, the contribution lies in four folds:

\begin{itemize}
    \setlength{\itemsep}{3pt}
    \setlength{\parsep}{3pt}
    \setlength{\parskip}{3pt}
    \item{We formally define the task of financial analysis, which aims to accomplish stock trend prediction and the corresponding financial Q\&A.
    }
    \item{We propose AlphaFin datasets, which contains traditional research datasets, real-time financial data, and handwritten CoT data, enhancing LLMs's ability in financial analysis.
    }
     \item {We fine-tune a StockGPT based on AlphaFin datasets and integrate it to a Stock-Chain framework, which is further integrated with a real-time financial database through RAG. 
     By integrating with RAG, we address the issue of the hallucination of LLMs's output and LLMs's inability to generate real-time content.
     }
     \item {We conduct extensive experiments on the AlphaFin datasets, to reveal that Stock-Chain outperforms all the baseline methods, and shows effectiveness for financial analysis.
     }
\end{itemize}

\begin{figure}[t]
\centering
\includegraphics[width=0.5\textwidth, keepaspectratio]{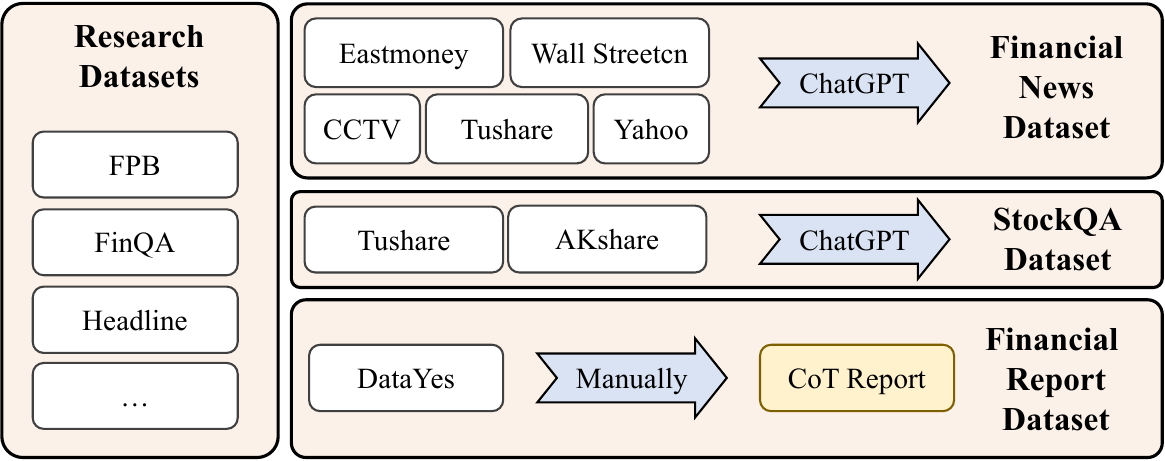}
\caption{The data source and preprocessing of the proposed AlphaFin datasets.
}
\label{fig2}
\end{figure}

\section{ AlphaFin Datasets }
We release AlphaFin datasets, as shown in Figure~\ref{fig2}, which include four parts, research datasets, StockQA, financial news, and financial reports. 
AlphaFin is sourced from a dozen of data sources. 
From Table~\ref{tab1}, it is evident that the traditional research dataset exhibits relatively shorter length of label, which hampers the training of FinLLMs. 
Thus, AlphaFin addresses the issue of low quality and length in traditional research datasets.
In this section, we provide the details of its sources and construction process.







\subsection{Data Sources}

\begin{table}[b] \small
\renewcommand\arraystretch{1.2}
\centering
\setlength{\tabcolsep}{4pt}{
\footnotesize
\begin{tabular}{l|c|c|c|c}
\toprule
 \textbf{Dataset}  &  \textbf{Size}  &  \textbf{Input} & \textbf{Label} & \textbf{Type} \\
\midrule
 Research  & 42,373 & 712.8 & 5.6 & en \\
 \hline
 StockQA  & 21,000 & 1313.6 & 40.8 & zh \\
 \hline
 Fin. News & 79,000 & 497.8 & 64.2 & zh \\
 \hline
 Fin. Reports  & 120,000 & 2203.0 & 17.2 & zh \\
 Fin. Reports CoT & 200 & 2184.8 & 407.8 & zh \\
\bottomrule
\end{tabular}
}
\normalsize
\linespread{1}
\caption{The details of the AlphaFin datasets. ``Input'' and ``Label'' denote their text length.}
\label{tab1}
\end{table}

\begin{itemize}
    \setlength{\itemsep}{3pt}
    \setlength{\parsep}{3pt}
    \setlength{\parskip}{3pt}
     \item {$\mathbf{Research\ datasets}$: This part includes traditional financial datasets from the academic, including FPB~\cite{malo2014good}, FinQA~\cite{maia201818}, convFinQA~\cite{chen2022convfinqa} and Headline~\cite{sinha2020impact}, etc. which enhance the information extraction and summarization ability for LLMs.}
    \item {$\mathbf{StockQA\ dataset}$: This part encompasses stock price and other financial data from Tushare~\cite{Tushare} and AKshare~\cite{AKshare}. It utilizes sequential data format, such as the real-world stock price trend  (e.g. $\{..., 170, 173, 171, 175, 173, 170, ...\}$).} 
    \item{$\mathbf{Financial\ News\ dataset}$: 
    To provide real-world financial knowledge for LLMs, we incorporate online news sources, such as the financial sections of CCTV, and Wall Street CN. }
    
    \item {$\mathbf{Financial\ reports\ dataset}$: We build financial report datasets via DataYes~\cite{DataYes}, including professional analysis and knowledge of companies conducted by institutions. 
     }
\end{itemize}
\vspace{-1em}

\subsection{Data Preprocessing}

As shown in Figure~\ref{fig2}, we explore the details of AlphaFin preprocessing: 

\begin{itemize}
    \setlength{\itemsep}{3pt}
    \setlength{\parsep}{3pt}
    \setlength{\parskip}{3pt}
    \item {$\mathbf{Research}$ $\mathbf{dataset}$: Traditional research datasets are primarily in English and of substantial quantity. To enhance the LLMs's ability in Chinese and ensure quality fine-tuning, we only sample a portion from the source.
    } 
    
    \item{$\mathbf{StockQA}$ $\mathbf{dataset}$:
    Given the source data is presented in sequential format, we utilize ChatGPT with the following prompt, to generate financial questions upon it.
    \texttt{Based on the ..., give me a good financial question. Input: <sequential data>, Output: <Question>.}
    This can facilitate training and enhance the diversity of questions. 
    Subsequently, we use ChatGPT to generate responses and obtain Q\&A pairs for training LLMs.
    }
    
    \item{$\mathbf{Financial}$ $\mathbf{News}$  $\mathbf{dataset}$: We leverage ChatGPT to extract a summary for each news, and construct the financial news dataset. This process improves LLMs's ability to generate summaries for financial news.
    }

     \item {$\mathbf{Financial}$ $\mathbf{reports}$ $\mathbf{dataset}$: 
    We manually align the financial reports for the companies and their stock price on the day of report publication, and use the following template to generate the final data. 
    \texttt{According to ... give a clear answer up or down. Input: <reports \& stock price>, Output: <Up/down>.} 
    
    Furthermore, we manually create 200 financial reports CoT data with professional financial knowledge and longer labels, to provide the LLMs with progressive analytical ability.
    The output format is: 
    
    \texttt{According to ... 
    conclusions can be drawn: 
    1. Fundamentals: ... 2. Technical aspects: ...\\Therefore, we predict the ... is <up/down>, probability: <Prob>}
     }
\end{itemize}
\vspace{-1em}

\section{Stock-Chain Framework}


We treat the financial analysis task as two counterparts, stock trend prediction and the corresponding financial Q\&A. Thus, our proposed Stock-Chain framework is divided into two stages as shown in Figure~\ref{fig3}.
In this section, we first formalize the task, and then introduce the details of both two stages.


\subsection{ Problem Definition }

For the first stage, given a set of companies $C = \{c_i\}_{i=1}^N$
and the corresponding knowledge documents $D = \{d_j\}_{j=1}^M$ 
, we can predict stock trends:\vspace{-6pt}

\begin{equation}
    Pred_i = \phi(c_i, d_j),\ \ Pred_i \in \{up, down\}\vspace{-3pt}
\end{equation}
where $\phi$ represents a stock predicting system, and $d_j$ is retrieved as the related documents for company $c_i$. The goal is to choose a subset of companies $C_{chosen}$ that are predicted to rise.\vspace{-6pt}

\begin{equation}
    C_{chosen} = \{c_i | c_i \in C \land Pred_i = up\}\vspace{-3pt}
\end{equation}

For the second stage, we treat a multi-turn dialogue session as a sequence of several query-response pairs between two interlocutors. We denote $Q_t$ and $R_t$ as the user query and agent response at the current time step $t$, and $H_t = [Q_0, R_0, ..., Q_{t-1}, R_{t-1}]$ as the dialogue history. Then we formalize the financial Q\&A task as obtaining the response based on the current query, dialogue history, and corresponding documents: \vspace{-6pt}
\begin{equation}
    R_t = \pi(d_k, H_t, Q_t)\vspace{-3pt}
\end{equation}
where $\pi$ represents the conversation system, and $d_k$ is the retrieved document related to $Q_t$.

\subsection{ Stage-1: Stock Trend Prediction }

\begin{figure*}[ht]
\centering
\includegraphics[width=1\textwidth, keepaspectratio]{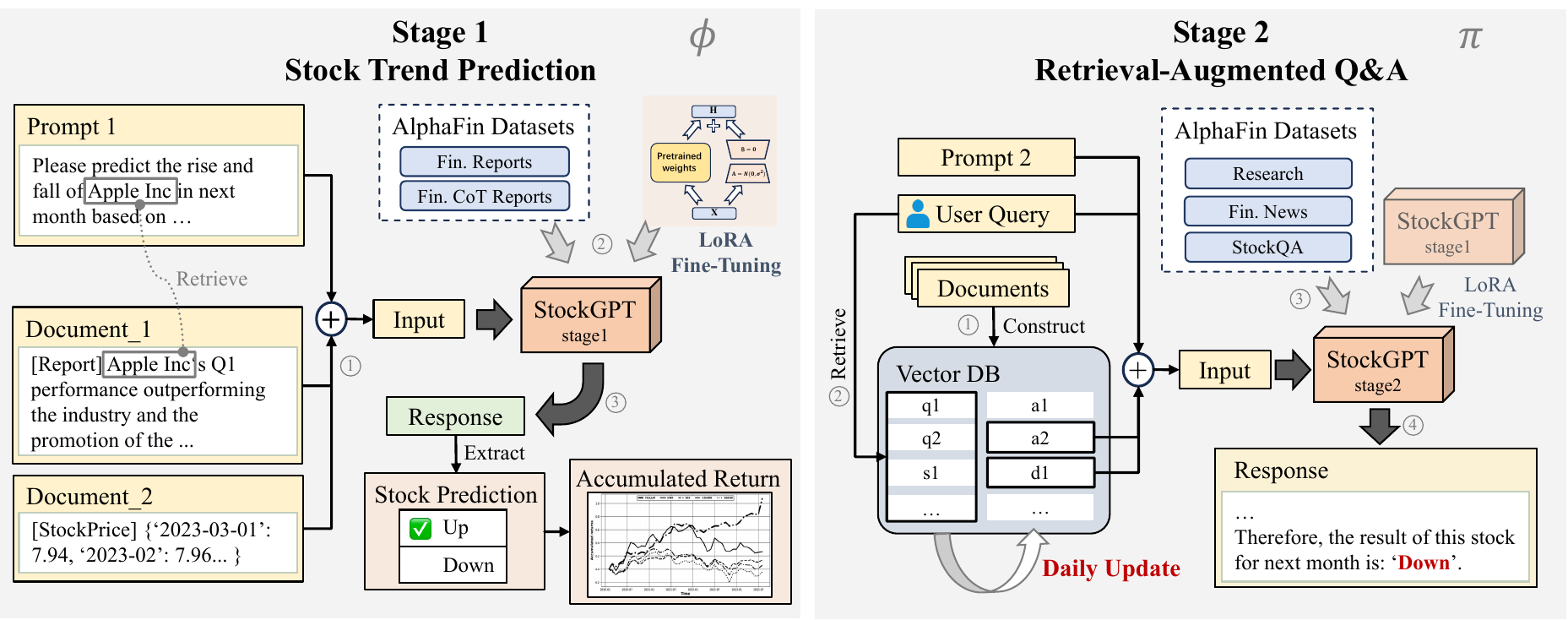}
\caption{An illustration of the Stock-Chain framework of the two stages in financial analysis.}
\label{fig3}
\end{figure*}

As shown in the left part of Figure~\ref{fig3}, our first stage is stock trend prediction. Given a company $c_i$ and the corresponding document $d_j$, this stage maintains a stock predicting system $\phi$ by combining LLMs and AlphaFin datasets, to give the stock trend prediction $Pred_i$ for $c_i$.



\subsubsection{Knowledge Processing}

As shown in Figure~\ref{fig3} \textcircled{1}, given a company $c_i$, we first retrieve the related documents $d_j$ for it.
Then, we design a prompt template $Prompt_1$ as follows:

\texttt{Please predict the rise and fall of the stock next month based on the research reports and data provided below. Please provide a clear answer, either ``up" or ``down". <report><market data>}

where \texttt{<report>} and \texttt{<market data>} compose $d_j$.
Ultimately, we concatenate the prompt with the documents to get the input $I_i$ to the LLMs.\vspace{-6pt}

\begin{equation}
    I_i = concat(Prompt_1, d_j)\vspace{-3pt}
\end{equation}

\subsubsection{StockGPT Fine-Tuning}




 As shown in Figure~\ref{fig3} \textcircled{2}, we design the fine-tuning process of an LLM, called StockGPT, which includes two steps. Firstly, we leverage all the financial report datasets of AlphaFin for training. In the second step, we utilize the manually created report CoT dataset to guide the model to think step-by-step. All fine-tuning processes for StockGPT utilize the Low-Rank Adaptation (LoRA)~\cite{hu2021lora} method.




Through the fine-tuning of the two steps, we obtain $StockGPT_{stage1}$, which is able to predict the trend of $c_i$ based on $d_j$ more accurately, as well as providing detailed analysis and explanations.

\subsubsection{Prediction and Post-process}

As shown in Figure~\ref{fig3} \textcircled{3}, given the Input $I_i$, we leverage StockGPT to predict the rise and fall of the stock, which can be viewed as a binary classification task.
By feeding $StockGPT_{stage1}$ the input $I_i$, we obtain the response text $Res_i$ about $c_i$.

The format of $Res_i$ could be referred to as Sec 2.2. The \texttt{<Prob>} is a category in the range of: $\{very\ large,\ large,\ medium\ to\ upper,\ average\}$, which could be a piece of supplementary information for the investor's decision-making.\vspace{-6pt}

\begin{equation}
    Res_i = StockGPT_{stage1}(I_i) \label{eq_res_stage1}\vspace{-3pt}
\end{equation}

Then, we manually extract the prediction result $Pred_i$ from $Res_i$.
Finally, we choose all stocks predicted as ``up'' as $C_{chosen}$.
\begin{equation}
    Pred_i = \left\{ 
        \begin{array}{ll}
            up, & if\ \ ``up" \in Res_i \\
            down, & else \\
        \end{array}
    \right.
\end{equation}

\begin{equation}
    C_{chosen} = \{c_i | Pred_i=``up"\}\vspace{-3pt}
\end{equation}

Additionally, we implement this investment strategy rolling by months. Every month, for all the $c_i$ in $C_{chosen}$, we hold them throughout the month.
The proportion of each stock in the portfolio is calculated via a capitalization-weighted approach.\vspace{-6pt}

\begin{equation}
    AR_m = AR_{m-1} + \sum_{c_i\in C_{chosen}}w* R_{c_i}\vspace{-3pt}
\end{equation}
where $AR_m$ is the accumulated return of month $m$, and $R_{c_i}$ is the return of stock $c_i$. $w$ denotes the proportion of stock $c_i$ in the portfolio. $v_i$ is the market value of company $c_i$.\vspace{-6pt}

\begin{equation}
    w = \frac{v_i}{\sum_{n=1}^Nv_n}\vspace{-3pt}
\end{equation}

\subsection{Stage-2: Financial Q\&A}

Besides stock trend prediction, the proposed Stock-Chain also has the ability for financial Q\&A, which could be more constructive for investors.

Given a dialogue history $H_t$, user query $Q_t$, and the retrieved document $d_j$ related to $Q_t$, conversation system $\pi$ can give a response $R_t$.
We adopt RAG to enhance the Q\&A ability of LLMs, which typically includes three parts: vector DB construction, knowledge retrieval, and response generation.


\subsubsection{Vector DB Construction}

As shown in Figure~\ref{fig3} stage 2 \textcircled{1}, vector DB is an important part of RAG, which is used for efficient storage and retrieval of knowledge documents.

\paragraph{Knowledge Extraction} To improve the accuracy and efficiency of document retrieval, we extract the key knowledge from the document. We adopt two extraction strategies: coarse-grained document-level summarizing with ChatGPT, and fine-grained entity-level dialogue generation through RefGPT~\cite{refgpt}. For document $d_k$, the extraction process for the two strategies is as follows:\vspace{-6pt}

\begin{equation}
    s_k = ChatGPT(d_k)\vspace{-6pt}
\end{equation}

\begin{equation}
    (q_{k0},a_{k0}), (q_{k1},a_{k1}), ... = RefGPT(d_k)\vspace{-3pt}
\end{equation}
where $s_k$ denotes the summary of $d_k$, and $(q_{k\_},a_{k\_})$ is query-answer pair of the generated dialogues. For example, for $d_k$ about k line, $q_{k\_}$ could be ``What is the meaning of k line?".

\paragraph{Knowledge Embedding} We take the extraction strategy of ChatGPT as an example. Given summary $s_k$, we obtain the embedding vector $e_{s_k}$ via a sentence embedding model. This vector would be stored in the database for subsequent retrieval.\vspace{-6pt}

\begin{equation}
    e_{s_k} = SentEmbed(s_k)\vspace{-3pt}
\end{equation}
where $SentEmbed$ is a sentence embedding model, such as BGE~\cite{bge_embedding} and SGPT~\cite{muennighoff2022sgpt}. We adopt BGE as the embedding model in our framework.

\paragraph{Continuous Updating}


Finally, we construct a vector DB including reports, market data, and financial news. The knowledge documents in the DB could be continuously updated via online data backflow, to keep the knowledge in real time.


\subsubsection{Knowledge Retrieval}

To retrieve knowledge in the vector DB, user query $Q$ would also be fed into the 
same sentence embedding model to obtain the embedding vector $e_Q$.\vspace{-6pt}

\begin{equation}
    e_Q = SentEmbed(Q)\vspace{-3pt}
\end{equation}

We choose the document with the highest cosine similarity to the query as the external knowledge aided StockGPT in generating responses.\vspace{-6pt}
\begin{equation}
    d^* = \argmax_{d_k} \frac{e_Q^{\top} \cdot e_{s_k}}{|e_Q||e_{s_k}|}\vspace{-3pt}
\end{equation}
where $s_k$ and $d_k$ could be replaced by $q_k\_$ and $a_k\_$ for the extraction strategy of RefGPT.

\subsubsection{LLMs Fine-Tuning}

We inherit $StockGPT_{stage1}$ as the base LLMs in this part, then continue training $StockGPT_{stage1}$ on the research dataset, financial news, and StockQA datasets of AlphaFin to obtain $StockGPT_{stage2}$.


\subsubsection{Response Generation}
Given a dialogue history $H_t$, user query $Q_t$, and retrieved document $d^*$ related to $Q_t$, the goal is to give the response $R_t$ in a conversation of turn $t$.
We provide a prompt template $Prompt_2$ as follows:


\texttt{You are an intelligent assistant, please answer my question. To help you ... local knowledge base is provided as follows: <knowledge>\\Now, answer the question...: <history><query>
}

Then, we concatenate the prompt template, retrieved knowledge, conversation history, and user query to get the input $I_t$ for LLMs. Feeding $I_t$ into StockGPT, we can get the response $R_t$.\vspace{-6pt}
\begin{equation}
    I_t = concat(Prompt_2, d^*, H_t, Q_t)\vspace{-3pt}
\end{equation}

\begin{equation}
    R_t = StockGPT_{stage2}(I_t)\vspace{-3pt}
\end{equation}

\section{ Experiments }

\begin{figure*}[ht]
\centering
\includegraphics[width=0.95\textwidth, keepaspectratio]{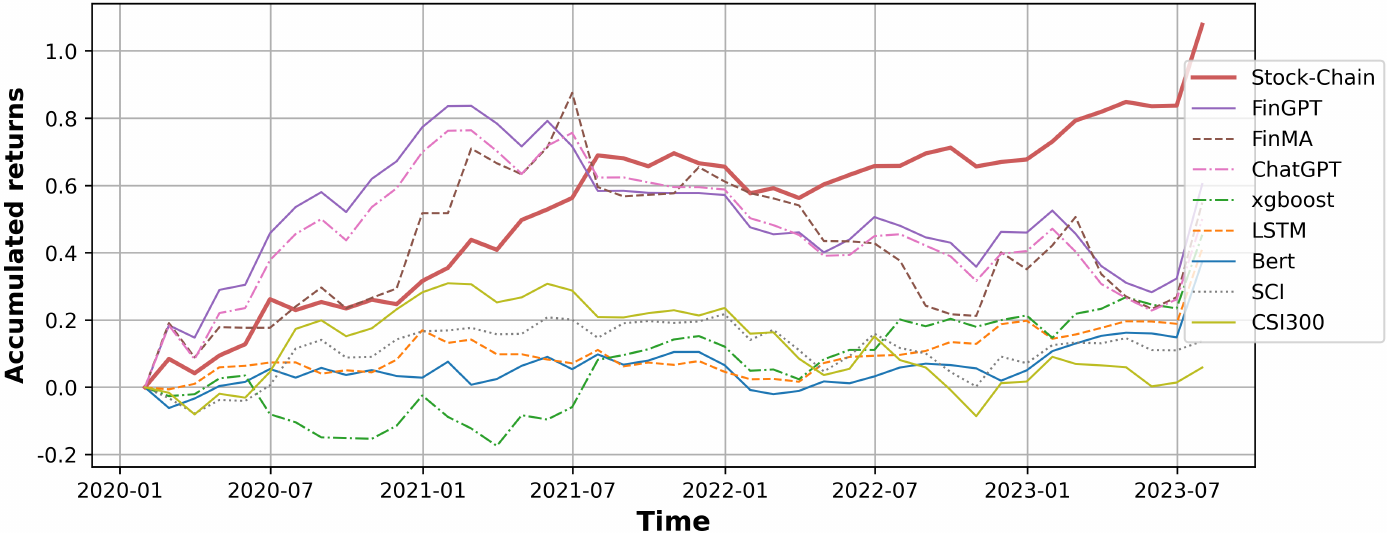}
\caption{Accumulated returns (AR) of each baseline under the test set of the financial report dataset from January 2020 to July 2023. The figure shows the curves of some baselines.
}
\label{fig4}
\end{figure*}

\begin{table*}[htb]\small
\renewcommand\arraystretch{1.2}
\centering
\setlength{\tabcolsep}{5pt}{
\footnotesize
\begin{tabular}{lc|cccccc|c}
\toprule
 \textbf{Model}  &  \textbf{ARR} $\uparrow$ & \textbf{AERR} $\uparrow$ & \textbf{ANVOL} $\downarrow$ & \textbf{SR} $\uparrow$ & \textbf{MD} $\downarrow$ & \textbf{CR} $\uparrow$ & \textbf{MDD} $\downarrow$ & \textbf{ACC} $\uparrow$ \\
\midrule
SSE50  & -1.0\% & -2.7\%  & 19.3\% & -0.054 & 45.9\% & -0.023 & 29 & - \\
CSI 300  & 1.7\% & 0 & 18.2\% & 0.092  &  39.5\% & 0.043 & 30 & - \\
SCI  & 3.9\% & 2.2\% & 14.8\% & 0.266  & 21.5\%  & 0.183 & 19 & - \\
CNX  & 7.6\% & 5.9\%  & 26.5\% & 0.287 & 41.3\% & 0.185 & 20 & - \\
\hline
\hline
Randomforest  & 9.8\% & 8.1\% & 19.5\% & 0.501 & 16\% & 0.608 & 22 & 55.5\% \\
RNN  & 8.1\% & 6.4\% & 10.9\% & 0.742 & 15.7\% & 0.515 & 12 & 54.1\% \\
BERT  & 10.7\% & 9.0\% & 16.1\% & 0.664  & 13.5\%  & 0.852 & 14 & 51.4\%  \\
GRU  & 11.2\% & 9.5\% & 13.7\% & 0.814 & 14.6\% & 0.765 & 21 & 54.7\% \\
LSTM  & 11.8\% & 10.1\% & 15.4\% & 0.767 & 15.3\%  & 0.768 & 19 & 55.2\% \\
Logistic  & 12.5\% & 10.8\% & 27.1\% & 0.463 & 32.5\% & 0.385 & 18 & 54.8\% \\
XGBoost  & 13.1\% & 11.4\% & 20.5\% & 0.633 & 20.9\% & 0.619 & 17 & \textbf{55.9\%} \\
Decision Tree  & 13.4\% & 11.7\% & 19.6\% & 0.683 & \textbf{11.9\%} & 1.126 & 20 & 55.1\% \\
\hline
\hline
ChatGLM2  & 8.1\% & 6.4\% & 24.9\% & 0.324  & 62.6\%  & 0.126 & 26 & 49.5\%  \\
ChatGPT(3.5Turbo)  & 14.3\% & 12.6\% & 27.7\% & 0.516 & 53.6\% & 0.267 & 23 & 51.4\% \\
FinMa  & 15.7\% & 14.0\% & 37.1\% & 0.422 & 66.3\% & 0.236 & 25 & 49.1\% \\
FinGPT  & 17.5\% & 15.8\% & 28.9\% & 0.605 & 55.5\% & 0.312 & 24 & 50.5\% \\
\hline
\hline
\textbf{Stock-Chain}  & \textbf{30.8\%} & \textbf{29.1\%} & \textbf{19.6\%}  & \textbf{1.573}  & 13.3\%  & \textbf{2.314}  & \textbf{10}  & 55.7\%  \\
\bottomrule
\end{tabular}
}
\normalsize
\linespread{1}
\caption{The main experimental results on AlphaFin-Test. 
ARR (Annualized rate of return) and ACC (Accuracy rate) are core indicators, while the middle indicators (like AERR, ANVOL, etc.) could assist investors in evaluating the model's effectiveness. Since the rate of return usually fluctuates wildly, to ensure the stability of the performance, we run each model 10 times and obtain the average result.
}
\vspace{-1em}
\label{tab2}
\end{table*}

In this section, we conduct experiments to validate Stock-Chain's ability to accomplish the task of financial analysis. Due to the structure of our framework, experiments can be divided into two parts. 
The experiments in the first part mainly examine the model's annualized rate of return and accuracy. 
In the second part, we demonstrate the performance of our Stock-Chain via preference evaluation with human\&GPT-4, ablation study, and case study.

\subsection{AlphaFin-Test Datasets}


We select a subset of data from the data sources, which is excluded from the training dataset, to serve as our test dataset.
Given that all the research datasets are in English, our main focus is on sampling from other datasets. Such as financial reports and StockQA datasets.
For stage 1, we choose the test datasets from the financial report dataset. An example demonstration is as follows:
\texttt{According to ..., please judge the trend of the company and give a clear answer up or down. Input: <reports \& stock price>, Output: <Up/down>.} 
As for stage 2, the test dataset is sampled from the StockQA and research datasets.
The AlphaFin-test datasets allow us to evaluate the model's ability in the capital market.


\subsection{Baselines}

To fully validate the effectiveness of our Stock-Chain on the test dataset, we select four categories of models:
\begin{itemize}
    \setlength{\itemsep}{3pt}
    \setlength{\parsep}{3pt}
    \setlength{\parskip}{3pt}
    \item {$\mathbf{Major}$ $\mathbf{Indices}$: We select indices in the Chinese capital market, including the SCI, CSI 300, SSE50, and CNX.}
    \item {$\mathbf{ML\&DL}$ $\mathbf{Algorithms}$: We employ ML algorithms such as Logistic and XGBoost, and DL models like LSTM and GRU, which are widely employed for time-series prediction.}
    \item {$\mathbf{General}$ $\mathbf{LLMs}$: We focus on the general-purpose LLMs like ChatGLM2 and ChatGPT. These LLMs have been chosen due to their ability and wide range of applications in NLP. }
    \item {$\mathbf{FinLLMs}$: In the financial domain, we focus on open-source FinLLM, such as FinGPT and FinMA, which have been trained for financial tasks like financial analysis and forecasting.}
\end{itemize} 

\subsection{Settings}

For stage 1, the experiments aim to predict the trend of stock price for the next month and observe the model's returns in the real market.
\begin{itemize}
    \setlength{\itemsep}{3pt}
    \setlength{\parsep}{3pt}
    \setlength{\parskip}{3pt}
    \item ML\&DL: Due to their limitations, they can only analyze time series data. Thus, their inputs are restricted to the stock prices.
    \item LLMs: In contrast, LLMs possess generative abilities, allowing them to incorporate both the report data and the stock price series data. 
\end{itemize}

For stage 2, we examine the model generation capability and use GPT4\&human as the evaluator. All LLMs' generation strategies are greedy search to achieve optimal and stable performance and integrated RAG in our experiments.

Among them, the hyperparameters are as follows: batch size 16, LoRA rank 8, cosine lr scheduler, learning rate 5e-5, bf16, and 1 NVIDIA A800-80GB for all training processes. Specifically, in stage 1, we trained 4 epochs for the first step and 20 epochs for the second step. In stage 2, we used $StockGPT_{stage1}$ as the base model, and incremental fine-tuned it for 2 epochs on the AlphaFin dataset.

\subsection{Metrics}

For stage 1, we utilize two categories of indicators. The first category is core indicators, including ARR and ACC, which gauge profitability. The second category is supplementary indicators that assist in analyzing different models, such as maximum drawdown (MD), Calmar Ratio (CR), and Sharpe ratio (SR), which measure risk assessment.
With these indicators, we have a thorough evaluation of the model's capabilities. 
For stage 2, we use the ROUGE~\cite{chin2004rouge} as an evaluation metric, which is used to measure the similarity between the generated output and reference information. Also, we use GPT4\&human as the referee for scoring.
By considering these indicators, we can better evaluate the performance of the models.

\begin{table}[htb]\small
\renewcommand\arraystretch{1.4}
\centering
\setlength{\tabcolsep}{5pt}{
\footnotesize
\begin{tabular}{lccccc}
\toprule
 \textbf{Model}  &  \textbf{ARR} $\uparrow$  &  \textbf{SR} $\uparrow$ &  \textbf{Out\_len} $\uparrow$ & \textbf{N/A} $\downarrow$ \\
\midrule
ChatGLM2  & 8.1\% & 0.324 & 228.1 & 52.3\% \\
\hline
w/ raw\_data & 15.8\% & 0.636 & 17.2 & \bf{-} \\
\hline
w/ CoT\_data & 10.1\% & 0.469 & \bf{476.1} & 32.4\% \\
\hline
\bf{Stock-Chain} & \bf{30.8\%} & \bf{1.573} & 254.8 & 25.9\%\\
\bottomrule
\end{tabular}
}
\normalsize
\linespread{1}
\caption{The ablation results for stage 1 under different training data. Out\_len: average length of LLMs outputs. N/A: invalid answer ratio.}
\label{tab3}
\end{table}

\begin{table}[htb]\small
\renewcommand\arraystretch{1.3}
\centering
\setlength{\tabcolsep}{1.0pt}{
\footnotesize
\begin{tabular}{lccccc}
\toprule
 \textbf{Model}  &  \textbf{ROUGE-1} $\uparrow$ & \textbf{ROUGE-2} $\uparrow$ & \textbf{ROUGE-L} $\uparrow$ \\
\midrule
ChatGLM2  & 0.2794 & 0.1944 & 0.2642 \\
\hline
w/ Fin. News & 0.3477 & 0.2821 & 0.3445 \\
\hline
w/ Fin. reports & 0.2611 & 0.1603 & 0.2396 \\
\hline
\bf{Stock-Chain} & \bf{0.4352} & \bf{0.3056} & \bf{0.4031} \\
\bottomrule
\end{tabular}
}
\normalsize
\linespread{1}
\caption{The ablation results for stage 2 under different training data. }
\vspace{-1em}
\label{tab4}
\end{table}

\subsection{Comparison Results}

As shown in Figure~\ref{fig4}, the curve represents the AR of each method. It is noteworthy that Stock-Chain achieves the highest AR and maintains an upward trend, starting from 2023. It indicates the effectiveness of Stock-Chain in investment. 

By referring to Table~\ref{tab2}, Stock-Chain achieves the highest 30.8\% of ARR demonstrating its effectiveness. 
Based on Table~\ref{tab2}, we can derive the following observations: 

Firstly, ML\&DL possesses certain analytical abilities in stock trend prediction, they achieve impressive ARR.
Secondly, after integrating report data with market data, LLMs generally surpass ML\&DL, leading to an enhancement in stock trend prediction abilities. ChatGPT achieves an ARR of 14.3\%. 
While LLMs are trained on vast amounts of textual data, they lack optimization for financial domains. 
Thus, by fine-tuning for financial domains, FinLLMs can improve stock trend prediction ability. 
The FinGPT model achieves an ARR of 17.5\%. 

Finally, after fine-tuning Stock-Chain based on financial report cot data, we achieve an ARR of 30.8\% and an ACC of 55.63\%. 
AlphaFin datasets play a crucial role in the training of LLMs. 
By utilizing comprehensive financial data for fine-tuning, we improve prediction accuracy and return, thus validating the performance of Stock-Chain.



\begin{figure}[ht]
\centering
\includegraphics[width=0.5\textwidth, keepaspectratio]{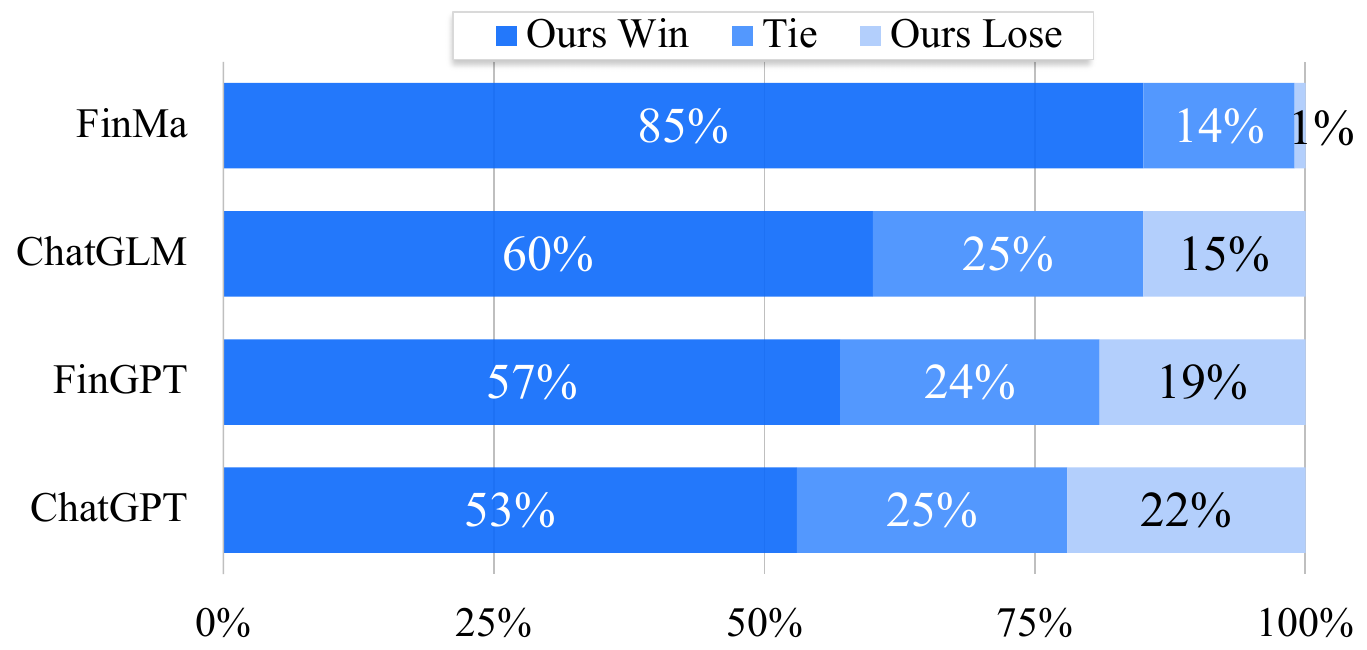}
\caption{Preference evaluations via human.}
\label{fig5}
\end{figure}

\begin{figure}[ht]
\centering
\includegraphics[width=0.5\textwidth, keepaspectratio]{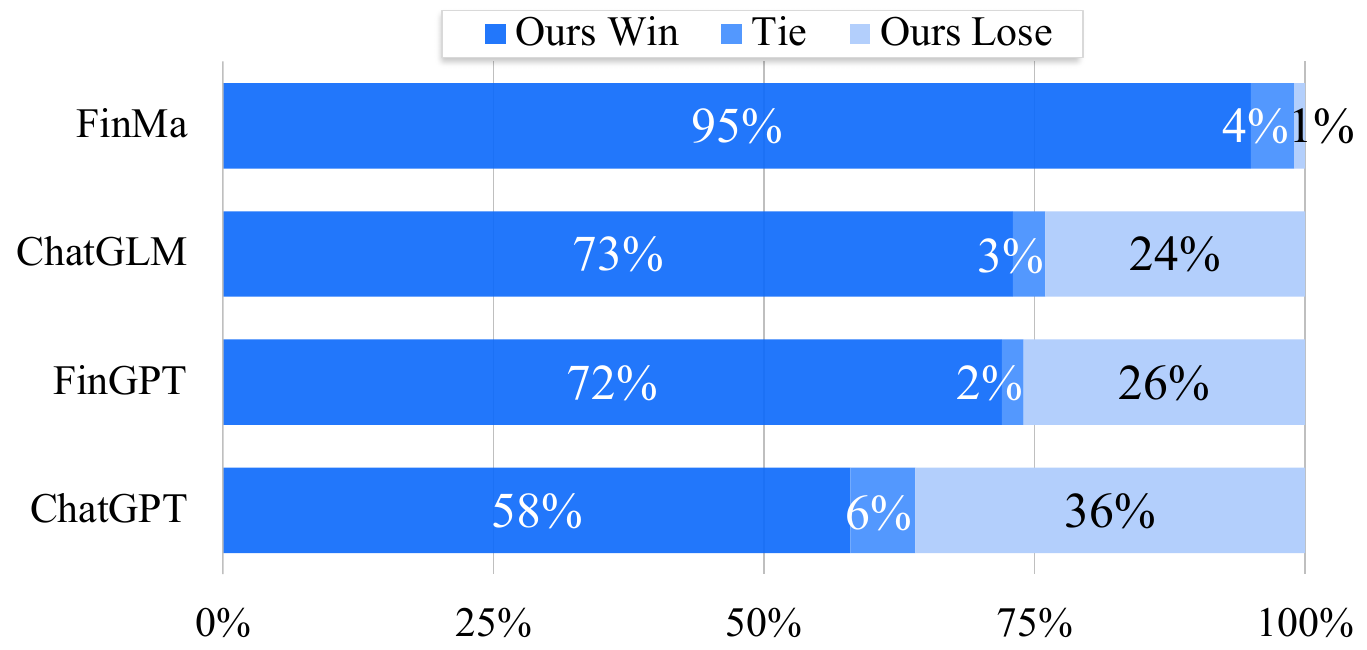}
\caption{Preference evaluations via GPT-4.}
\vspace{-1em}
\label{fig6}
\end{figure}

\begin{figure*}[ht]
\centering
\includegraphics[width=1.0\textwidth, keepaspectratio]{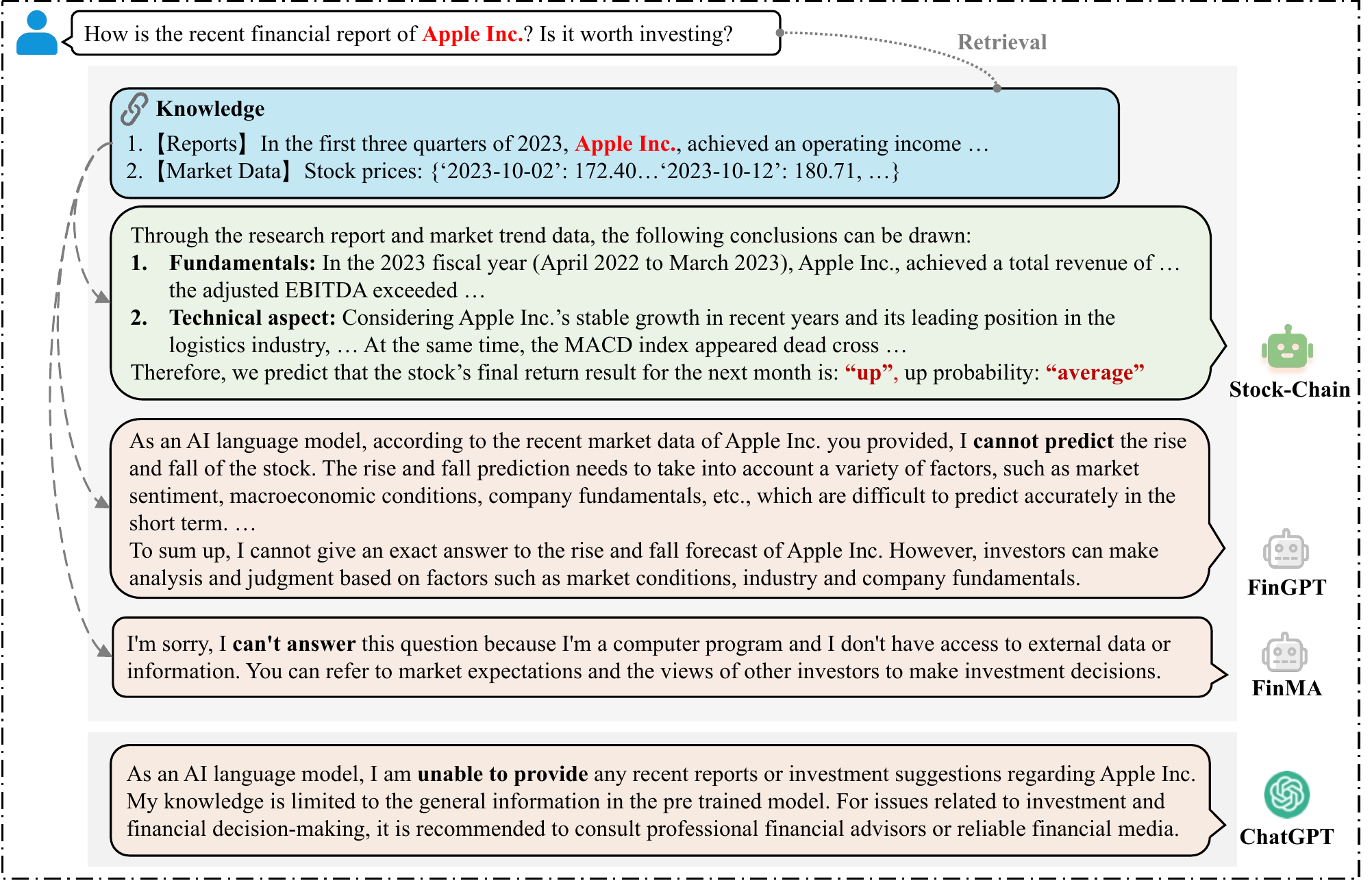}
\caption{Case of outputs of Stock-Chain, FinGPT, FinMA, and ChatGPT}
\vspace{2em}
\label{fig7}
\end{figure*}


\subsection{Ablation Study}

We conduct two ablation experiments.
Firstly, we observe the stock trend prediction ability of our Stock-Chain by analyzing the effects of fine-tuning at different data.
Based on Table~\ref{tab3}, compared to the ChatGLM2, the ability of the LLMs to predict stock price has shown improvement after fine-tuning with raw and CoT data, achieving returns of 15.8\% and 10.1\% respectively.

Additionally, the proportion of invalid answers also improved. 
It is worth mentioning that after fine-tuning with the raw data, the LLMs's output only includes rise and fall, thus resolving the issue of invalid answers. 
After fine-tuning with two sets of data, our Stock-Chain achieves optimal performance with 30.8\% ARR, and the proportion of invalid answers also decreased, reaching 25.9\%.

As for the second ablation experiment, we investigate whether the quality of the output improved after fine-tuning the LLMs at different data.

Based on Table~\ref{tab4}, we observe that the scores of Stock-Chain in terms of rouge1 and rouge2 reach 0.3477 and 0.2821 respectively after fine-tuning with News data. 
Furthermore, it is noteworthy that Stock-Chain achieves optimal performance after fine-tuning with both news and reports. 

\subsection{Preference Evaluation}

We employ GPT-4 and humans as the judge to rate the output performance of each LLM on the test datasets. All LLMs have integrated RAG in this experiment.
In the human section, Stock-Chain outperforms other LLMs in terms of content effectiveness. Based on Figure~\ref{fig5}, compared to ChatGLM2, Stock-Chain achieves a win rate of over 60\%, and when compared to FinLLMs, such as FinGPT, the win rate reaches 62\%. 
Based on Figure~\ref{fig6}, when GPT4 is the judge, similar conclusions were drawn. Stock-Chain exhibits higher rates compared to human ratings, with a win rate of 58\% against ChatGPT and 73\% against ChatGLM2. 
Overall, Stock-Chain's output is effective.

\subsection{Case Study}

 We present partial outputs of Stock-Chain for qualitative analysis. As shown in Figure~\ref{fig7}, when users inquire about recent reports and investment advice related to Appe Inc., Stock-Chain acquires real-time relevant information and market data from the knowledge base and provides them as input to the LLMs. The output of the LLMs has been enhanced, and the news remains up-to-date, enabling investors to analyze and receive recommendations. 
However, as for ChatGPT and FinGPT, we observe that it has great defects in the quality and real-time response.
Thus, by integrating RAG, Stock-Chain addresses the issues of hallucinations and insufficient real-time outputs in LLM, enhancing the LLMs's practicality and ability.

\section{Related Work}
\label{sec:append-how-prod}
\subsection{Financial Datasets}

General financial datasets include a wealth of information derived from various sources within the financial industry, such as Internet data and proprietary data. Currently, the main financial dataset sets include a variety of tasks. FPB~\cite{malo2014good} and FiQA-SA~\cite{maia201818} are mainly used for emotion analysis. The Headline~\cite{sinha2021impact} dataset is primarily utilized for news headline identification. For question answering, we primarily utilize the FinQA~\cite{chen2022finqa} and ConvFinQA~\cite{chen2022convfinqa} datasets.
Regrettably, the finance domain still suffers from a scarcity of text datasets, impeding the development of FinLLMs.
To bridge this gap, we propose AlphaFin, providing support for the financial industry in training its own FinLLMs.




\subsection{Algorithms in Financial Domain}

Traditional ML\&DL algorithms, such as LSTM~\cite{yu2019review}, Logistic~\cite{sperandei2014understanding}, and BERT~\cite{devlin2018bert}, have been applied in stock trend prediction. However, ML\&DL focuses on the final result, without analyzing the underlying factors driving market trend.
As for FinLLMs, although BloombergGPT~\cite{wu2023bloomberggpt}, FinMA~\cite{xie2023pixiu}, and FinGPT~\cite{yang2023fingpt} play important roles in the community, they are mainly based on English-language datasets. In contrast, Stock-Chain relies on Chinese-language and is specifically designed for stock trend prediction.

As for RAG, it has attracted growing attention in the community~\cite{li2022survey}. Compared to the traditional method, RAG has remarkable performance in various NLP tasks~\cite{cai2021neural,weston2018retrieve}. 
However, without RAG, FinLLMs often produce hallucinations and meaningless outputs. Thus, we integrate LLMs with RAG to address the above issues.

\section{Conclusion}

In this work, we formalize the task of financial analysis, and propose AlphaFin datasets to enhance LLMs's ability, and fine-tune the StockGPT upon it. Then we propose the Stock-Chain framework that integrates a real-time financial database through RAG, to address the issue of hallucination of LLMs's output and LLMs's inability to generate real-time content. We conduct extensive experiments on the proposed AlphaFin datasets, as well as some supplementary experiments like ablation study, GPT4\&human preference evaluation and case study, to reveal that Stock-Chain outperforms all the baseline methods, and shows effectiveness for the task of financial analysis.

\section{Ethical considerations and limitations}

We assert that there are no ethical dilemmas surrounding the submission of this article and have no known competing financial interests or personal relationships that could have had an impact on the research work presented.

Despite the positive contributions of this study, we recognize that there is still great room for development in our work. In our future work, we will further contribute to the open-source FinLLMs, improve its generalization, enhance its ability in other financial tasks, and create a more powerful open-source FinLLMs.

\section{Bibliographical References}\label{sec:reference}

\bibliographystyle{lrec-coling2024-natbib}
\bibliography{lrec-coling2024-example}

\end{document}